\documentclass{llncs}
\usepackage[utf8]{inputenc}
\usepackage{graphicx}
\usepackage{amsmath}
\usepackage{amssymb}
\usepackage{mathtools}
\usepackage{multirow}
\usepackage{stmaryrd}
\usepackage{multirow}
\usepackage{etoolbox}
\usepackage{moresize}
\usepackage{placeins}

\usepackage[caption=false]{subfig}
\usepackage[hyphens]{url}
\usepackage[breaklinks]{hyperref}
\usepackage{booktabs} 
\usepackage[bottom]{footmisc}
\usepackage[ruled,linesnumbered]{algorithm2e}
  \SetAlFnt{\scriptsize\sf}
  \def\algAssign{:=}

\def\ptFiguresDirectory#1{./figures/#1}
\def\eucDist#1{\left\|{#1}\right\|}
\def\FWER#1{FWER}
\def\fdr{\mathrm{FDR}}
\def\fnr{\mathrm{FNR}}

\def\mcc{\mathrm{MCC}}

\def\tablenameM{Table}

\begin{document}
\frontmatter          
\pagestyle{headings}  
\addtocmark{} 
\mainmatter              
\title{Soft Confusion Matrix Classifier for Stream Classification}
\titlerunning{SCM for Stream Classification}  
%
\author{Pawel Trajdos\orcidID{0000-0002-4337-6847} \and Marek Kurzynski\orcidID{0000-0002-0401-2725}}
\authorrunning{P. Trajdos \and M. Kurzynski} 
%
\tocauthor{Pawel Trajdos, Marek Kurzynski}
\institute{Wroclaw University of Science and Technology, Wroclaw, Poland,\\
\email{pawel.trajdos@pwr.edu.pl}, \email{marek.kurzynski@pwr.edu.pl}}

\maketitle              

\begin{abstract}
In this paper, the issue of tailoring the soft confusion matrix (SCM) based classifier to deal with stream learning task is addressed. The main goal of the work is to develop a wrapping-classifier that allows incremental learning to classifiers that are unable to learn incrementally. The goal is achieved by making two improvements in the previously developed SCM classifier. The first one is aimed at reducing the computational cost of the SCM classifier. To do so, the definition of the fuzzy neighbourhood of an object is changed. The second one is aimed at effective dealing with the concept drift. This is done by employing the ADWIN-driven concept drift detector that is not only used to detect the drift but also to control the size of the neighbourhood. The obtained experimental results show that the proposed approach significantly outperforms the reference methods. 

\end{abstract}

\section{Introduction}
Classification of streaming data is one of the most difficult problems in modern pattern recognition theory and practice. This is due to the fact that a typical data stream is characterized by several features that significantly impede making the correct classification decision. These features include: continuous flow, huge data volume, rapid arrival rate, and susceptibility to change  \cite{Krawczyk2017}. 
If a streaming data classifier aspires to practical applications, it must face these requirements and have to satisfy numerous constraints (e.g. bounded memory, single-pass, real-time response, change of data concept) to an acceptable extent. It is not easy, that is why the methodology of recognizing stream data has been developing very intensively for over two decades, proposing new, more and more perfect classification methods~\cite{Gama,Mehta}.

Incremental learning is a vital capability for classifiers used in stream data classification~\cite{Read2012batch}.  It allows the classifier to utilize new objects generated by the stream to improve the model built so far. It also allows, to some extent, dealing with the concept drift.  Some of the well-known classifiers are naturally capable to be trained iteratively. Examples of such classifiers are neural networks, nearest neighbours classifiers, or probabilistic methods such as the naive Bayes classifier~\cite{giraud2000note}. Some of the classifiers were tailored to be learned incrementally. An example of such a method is well-known Hoeffding Tree classifier~\cite{Pfahringer2007}. Those types of classifiers can be easily used in stream classification systems. On the other hand, when a classifier is unable to learn in an incremental way, the options for using for stream classification are very limited~\cite{Read2012batch}. Only one option is to keep a set of objects and rebuild the classifier from scratch whenever it is necessary~\cite{giraud2000note}. 

To bridge this gap, we propose a wrapping-classifier-based on the soft confusion matrix approach (SCM). The wrapping-classifier may be used to add incremental learning functionality to any batch classifier. The classifier based on the idea of soft confusion matrix has been proposed in~\cite{AMCS}. It proved to be an efficient tool for solving such practical problems as hand gesture recognition~\cite{CBM}. An additional advantage in solving the above-mentioned classification problem is the ability to use imprecise feedback information about a class assignment. The SCM-based algorithm was also successfully used in multilabel learning~\cite{IJNS}. 

Dealing with the concept drift using incremental learning only is insufficient. This is because the incremental classifiers deal effectively only with the incremental drift~\cite{Gama}. To handle the sudden concept drift, additional mechanism such as single/multiple window approach~\cite{kuncheva2009window}, forgetting mechanisms~\cite{vzliobaite2011combining}, drift detectors~\cite{Bifet2007} must be used. In this study, we decided to use ADWIN algorithm~\cite{Bifet2007} to detect the drift and to manage the set of stored objects. We use the ADWIN-based detector because this approach was shown to be an effective method~\cite{Gonalves2014,Barros2019}.

The concept drift may also be dealt with using ensemble classifiers~\cite{Gama}. There are a plethora of ensemble-based approaches~\cite{Gomez,Brzezinski2014ReactingTD,Krawczyk2017} however, in this work we are focused on single-classifier-based systems.

The rest of the paper is organized as follows. Section~\ref{sec:ClassifierCorrection} presents  the corrected classifier and gives  insight into its two-level structure and the original concepts of RRC and SCM  which are the basis of its construction. Section~\ref{sec:DSClassification} describes the adopted model of concept drifting data stream and provides details of chunk-based learning scheme of base classifiers and online dynamic learning of the correcting procedure and describes the method of combining ensemble members. 
In section~\ref{sec:ExpSetup} the description of the experimental procedure is given. The results are presented and discussed in section~\ref{sec:ResAndDisc}. Section~\ref{sec:Conclusions} concludes the paper.

\section {Classifier with Correction}\label{sec:ClassifierCorrection}

\subsection{Preliminaries}\label{sec:ClassifierCorrection:Preliminaries}

Let us  consider the pattern recognition problem in which $x \in \mathcal{X}$ denotes a feature vector of an object and $j \in \mathcal{M}$ is its class number ($\mathcal{X}\subseteq \Re^{d}$ and $\mathcal{M}=\{1,2,\ldots,M\}$ are feature space and set of class numbers, respectively). 
Let $\psi(\mathcal{L})$ be a classifier trained on the learning set $\mathcal{L}$, which assigns a class number $i$ to the recognized object.
We assume that $\psi(\mathcal{L})$ is described by the canonical model \cite{Kuncheva}, i.e. for given $x$ it first produces values of normalized classification functions (supports) $g_i(x), i \in \mathcal{M}$ ($g_i(x) \in [0, 1], \sum g_i(x)=1$) and then classify object according to the maximum support rule:
\begin{equation}   \label{wzor1}
\psi(\mathcal{L},x)=i  \Leftrightarrow  g_i(x) = \max_{k \in \mathcal{M}} g_k(x).
\end{equation}

To recognize the object $x$ we will apply the original procedure, which using additional information about the local (relative to $x$) properties of $\psi(\mathcal{L})$ can change its decision to increase the chance of correct classification of $x$. 

The proposed correcting procedure which has the form of  classifier  $\psi^{(Corr)}(\mathcal{L},\mathcal{V})$  built over $\psi(\mathcal{L})$ will be called a wrapping-classifier. The wrapping classifier $\psi^{(Corr)}(\mathcal{L},\mathcal{V})$ acts according to the following Bayes scheme:
\begin{equation}   \label{wzor2}
\psi^{(Corr)}(\mathcal{L},\mathcal{V},x)=i \Leftrightarrow  P(i|x) = \max_{k \in \mathcal{M}} P(k|x), 
\end{equation}
where \textit{a posteriori} probabilities $P(k|x), k \in \mathcal{M}$ can be expressed in a form depending on the probabilistic properties of classifier $\psi(\mathcal{L})$:
\begin{equation}   \label{wzor3}
P(j|x)=\sum_{i \in \mathcal{M}} P(i,j|x) = \sum_{i \in \mathcal{M}} P(i|x) P(j|i,x).
\end{equation}
$P(j|i,x)$ denotes the probability that $x$ belongs to the $j$-th class given that $\psi(\mathcal{L},x)=i$ and   $P(i|x)=P(\psi(\mathcal{L},x)=i)$ is the probability of assigning $x$ to class $i$ by $\psi(\mathcal{L})$
Since for deterministic classifier $\psi(\mathcal{L})$ both above probabilities are equal to 0 or 1 we will use two concepts for their approximate calculation: randomized reference classifier (RRC) and soft confusion matrix (SCM).

\subsection{Randomized Reference Classifier (RRC)}\label{sec:ClassifierCorrection:RRC} 
RRC is a randomized model of classifier $\psi(\mathcal{L})$  and with its help the probabilities $P(\psi(\mathcal{L},x)=i)$ are calculated.

RRC $\psi^{RRC}$ as a probabilistic classifier is defined by a probability distribution over the set of class labels $\mathcal{M}$. Its classifying functions $\{\delta_j(x)\}_{j \in \mathcal{M}}$ are observed values of random variables $\{\Delta_j(x)\}_{j \in \mathcal{M}}$ that meet -- in addition to the normalizing conditions -- the following condition:
\begin{equation}  \label{wzor4}
\mathbf{E}\left[\Delta_{i}(x) \right] = g_{i}(x),\ i \in \mathcal{M},
\end{equation}
where $\mathbf{E}$ is the expected value operator. Formula (\ref{wzor4}) denotes that $\psi^{RRC}$ acts -- on average -- as the modeled classifier $\psi(\mathcal{L})$, hence the following 
approximation is fully justified:
\begin{equation}  \label{wzor5}
P(i|x)=P(\psi(\mathcal{L},x)=i)\approx P(\psi^{RRC}(x)=i),
\end{equation}
where
\begin{equation}  \label{wzor6}
P(\psi^{RRC}(x)=i)=P [\Delta_i(x) > \Delta_k(x), k \in \mathcal{M} \setminus i]
\end{equation}
can be easily determined if we assume -- as in the original work of Woloszynski and Kurzynski \cite{PR} -- that  $\Delta_i(x)$ follows the beta distribution.

\subsection{Soft Confusion Matrix (SCM)}\label{sec:ClassifierCorrection:SCM}

SCM will be used to determine the assessment of probability $P(j|i,x)$ which denotes class-dependent probabilities  of the correct classification (for $i=j$) and the misclassification (for $i \neq j$) of $\psi(\mathcal{L},x)$ at the point $x$. The method defines the neighborhood of the point $x$ containing validation objects in terms of fuzzy sets allowing for flexible selection of membership functions and  assigning weights to individual validation objects dependent on distance from $x$. 

The SCM providing an image of the classifier local (relative to $x$) probabilities $P(j|i,x)$, is in the form of two-dimensional table, in which the rows correspond to the true classes while the columns correspond to the outcomes of the classifier $\psi(\mathcal{L})$, as it is shown in Table 1.

\begin{table}
{
\renewcommand{\arraystretch}{0.9}%
\setlength\tabcolsep{.9pt}%
\caption{The soft confusion matrix of classifier $\psi(\mathcal{L})$}
\label{MK_PT:confmatrix}
\begin{center}
\begin{tabular}{cc|cccc}
& & \multicolumn{4}{c}{Classification by $\psi$}\\
& &  $1$ & $2$ & \ldots & $M$ \\
\midrule
& $1$& $\varepsilon_{1,1}(x)$&$\varepsilon_{1,2}(x)$& \ldots&$\varepsilon_{1,M}(x)$\\
True& $2$& $\varepsilon_{2,1}(x)$&$\varepsilon_{2,2}(x)$& \ldots&$\varepsilon_{2,M}(x)$\\
class& $\vdots$  & $\vdots$ & $\vdots$ & $\ddots$  &$\vdots$ \\
& $M$& $\varepsilon_{M,1}(x)$&$\varepsilon_{M,2}(x)$& \ldots&$\varepsilon_{M,M}(x)$\\
\end{tabular}
\end{center}
}
\end{table}
The value $\varepsilon_{i,j}(x)$ is determined from validation set $\mathcal{V}$ and is defined as the following ratio: 
 
\begin{equation}  \label{wzor7}
\varepsilon_{i,j}(x)= \frac{|\mathcal{V}_j \cap \mathcal{D}_i \cap \mathcal{N}(x)|}{|\mathcal{V}_j \cap  \mathcal{N}(x)|}, 
\end{equation}
where $\mathcal{V}_j, \mathcal{D}_i$ and $\mathcal{N}(x)$ are fuzzy sets specified in the validation set $\mathcal{V}$ and $|\cdot|$  denotes the cardinality of a~fuzzy set~\cite{Dhar2013}.

The set $\mathcal{V}_j$  denotes the set of validation objects from the $j$-th class. Formulating this set  in terms of fuzzy sets theory it can be assumed that the grade of membership of validation object $x_{\mathcal{V}}$ to  $\mathcal{V}_j$ is the class indicator which leads to the following definition of $\mathcal{V}_j$:

\begin{align} \label{wzor8}
\mathcal{V}_j&=\{(x_{\mathcal{V}}, \mu_{\mathcal{V}_j}(x_{\mathcal{V}}))\},\\
\mu_{\mathcal{V}_j}(x_{\mathcal{V}})&=\left\{\begin{array}{ll}
1 & \textrm{if}\ x_{\mathcal{V}} \in  j\textrm{-th class,}\\
0 & \textrm{elsewhere.}
\end{array} \right.
\end{align}

The concept of fuzzy set $\mathcal{D}_i$ is defined as follows:
\begin{equation}  \label{wzor9}
\mathcal{D}_i=\{(x_{\mathcal{V}}, \mu_{\mathcal{D}_i}(x_{\mathcal{V}})): x_{\mathcal{V}} \in \mathcal{V}, \mu_{\mathcal{D}_i}(x_{\mathcal{V}})=P(i|x_{\mathcal{V}})\},
\end{equation}
where $P(i|x_{\mathcal{V}})$ is calculated according to \eqref{wzor5} and \eqref{wzor6}. Formula \eqref{wzor9} demonstrates that the membership of validation object $x_{\mathcal{V}}$ to the set $\mathcal{D}_i$ is not determined by the decision of classifier $\psi(\mathcal{L})$. The grade of membership  of object $x_{\mathcal{V}}$ to  $\mathcal{D}_i$ depends on the potential chance of classifying  $x_{\mathcal{V}}$ to the $i$-th class by the classifier $\psi(\mathcal{L})$. We assume, that this potential chance is equal to the probability $P(i|x_{\mathcal{V}})=P(\psi(\mathcal{L},x_{\mathcal{V}})=i)$ calculated approximately using the randomized model RRC of classifier $\psi(\mathcal{L})$. 

Set $\mathcal{N}(x)$ plays the crucial role in the proposed concept of SCM, because it decides which validation objects $x_{\mathcal{V}}$ and with which weights will be involved in determining the local properties of the classifier $\psi(\mathcal{L})$ and -- as a consequence -- in the procedure of correcting its classifying decision. 
Formally, $\mathcal{N}(x)$ is also a fuzzy set:
\begin{equation} \label{wzor10}
\mathcal{N}(x)=\{(x_{\mathcal{V}}, \mu_{\mathcal{N}(x)}(x_{\mathcal{V}}))\},
\end{equation}
but its membership function is not defined univocally because it depends on many circumstances. By choosing the shape of the membership function $\mu_{\mathcal{N}(x)}$ we can freely model the adopted concept of "locality" (relative to $x$).

$\mu_{\mathcal{N}(x)}(x_{\mathcal{V}})$ depends on the distance between validation object $x_{\mathcal{V}}$ and test object $x$: its value is equal to 1 for $x_{\mathcal{V}}=x$ and decreases with increasing the distance between $x_{\mathcal{V}}$ and $x$. This leads to the following form of the proposed membership function of the set:
\begin{equation}  \label{wzor14a}
\mu_{\mathcal{N}}(x_{\mathcal{V}})= \left\{ \begin{array}{lcr} C \exp(-\beta \eucDist{x - x_{\mathcal{V}}}^2), & \mathrm{if} & \eucDist{ x - x_{\mathcal{V}}}<K_{d}\\
                                               0 &\mathrm{otherwise} &
                                              \end{array}\right.
\end{equation}
 $\eucDist{\cdot}$ denotes Euclidean distance in the feature space $\mathcal{X}$, $K_{d}$ is the Euclidean distance between $x$ and the $K$-th nearest neighbor in $\mathcal{V}$, $\beta \in \Re_+$ and $C$ is a normalizing coefficient. The first factor in \eqref{wzor14a} limits the concept of ``locality'' (relatively to $x$) to the set of $K$ nearest neighbors with Gaussian model of membership grade.
 
Since under the stream classification framework, there should be only one pass over the data~\cite{Krawczyk2017}, $K$ and $\beta$ parameters cannot be found using the extensive grid search approach just like it was for the originally proposed approach~\cite{AMCS,CBM}. Consequently, in this work, we decided to set $\beta$ to 1. Additionally, the initial number of nearest neighbours is found using a simple rule of thumb~\cite{Devroye1996}:
\begin{equation}
 \hat{K} = \left \lceil{\sqrt{|\mathcal{V}|}}\right \rceil. 
\end{equation}
To avoid ties, the final number of neighbours $K$ is set as follows: 
\begin{equation}
 K = \left\{ \begin{array}{lcr}
              \hat{K} & \mathrm{if} & M \mod \hat{K} \neq 0\\
              \hat{K}+1 & \mathrm{otherwise}&
             \end{array}
\right.
\end{equation}
Additionally, the computational cost of computing the neighbourhood may be further reduced by using the kd-tree algorithm to find the nearest neighbours~\cite{Hou2018}. 

Finally, from \eqref{wzor8}, \eqref{wzor9} and \eqref{wzor10} we get the following approximation:
\begin{equation}  \label{wzor11}
P(j|i,x) \approx \frac{\varepsilon_{i,j}(x)}{\sum_{j \in \mathcal{M}} \varepsilon_{i,j}(x)}, 
\end{equation}
which  together with \eqref{wzor3}, \eqref{wzor5} and \eqref{wzor6} give \eqref{wzor2} i.e. the corrected  classifier $\psi^{\mathrm{Corr}}(\mathcal{L}, \mathcal{V})$.

\subsection{Creating the validation set}\label{sec:ClassifierCorrection:ValSet}

In this section, the procedure of creating the validation set $\mathcal{V}$ from the training $\mathcal{L}$ is described. In the original work describing SCM~\cite{AMCS}, the set of labelled data was wplit into the learning set $\mathcal{L}$ and the validation set $\mathcal{V}$. The learning set and the validation set were disjoint $\mathcal{L}^\prime \cap \mathcal{V} =\emptyset$.  The cardinality of the validation set was controlled by the $\gamma$ parameter $|\mathcal{V}|=\gamma|\mathcal{L}|$, $\gamma \in [0,1]$. The $\gamma$ coefficient was usualy set to $0.6$, however to achieve the highest classification quality, it should be determined using the grid-search procedure. As it was said above, in this work we want to avoid using the grid-search procedure. Therefore, we construct the validation set using three-fold cross-validation procedure that allows using of the entire learning set as a validation set. The procedure is described in Algorithm~\ref{alg:SCM:Train}.

\begin{algorithm}[H]
 \KwData{
 $\mathcal{L}$ -- Initial learning set\;
 }
 \KwResult{
  $\mathcal{V}$ -- Validation set\;
  $\mathcal{D}_i$ -- Decision sets (see~\eqref{wzor9})\;
  $\psi(\mathcal{L})$ -- Trained classifier. 
 }
 \Begin{
 \For{$k \in \{1,2,3\}$}{
  Extract fold specific training and validation set $\mathcal{L}_k$, $\mathcal{V}_k$\;
  Learn the $\psi(\mathcal{L}_k)$ using $\mathcal{L}_k$\;
  $\mathcal{V} \algAssign  \mathcal{V} \cup \mathcal{V}_k$ \;
  Update the class-specific decision sets $\mathcal{D}_i$ using predictions of $\psi(\mathcal{L}_k)$ for instances from $\mathcal{V}_k$ (see~\eqref{wzor9})\;
}
 Learn the $\psi(\mathcal{L})$ using $\mathcal{L}$\;
}
 \caption{Procedure of training the SCM classifier. Including the procedure of validation set creation.\label{alg:SCM:Train}}
\end{algorithm}

\section{Classification of Data Stream}\label{sec:DSClassification}
The main goal of the work is to develop a wrapping-classifier that allows incremental learning to classifiers that are unable to learn incrementally. In this section, we describe the incremental learning procedure used by the SCM-based wrapping-classifier.

\subsection{Model of Data Stream}\label{sec:DSClassification:streamModel}

We assume that instances from a data stream $\mathcal{S}$ appear as a sequence of labeled examples  
$\{(x^t, j^t)\}, t=1,2,...,T$, where $x^t \in \mathcal{X} \subseteq \Re^d$ represents a $d$-dimensional feature vector of an object that arrived at time $t$ and $j^t \in \mathcal{M}=\{1,2,\ldots, M\}$ is its class number. In this study we consider a completely supervised learning approach which means that the true class number $j^t$ is available after the arrival of the object $x^t$ and before the arrival of the next object $x^{t+1}$ and this information may be used by classifier for classification of $x^{t+1}$. Such a framework is one of the most often considered in the related literature \cite{Brzez2014,Nguyen}.

In addition, we assume that a data stream can be generated with a time-varying distribution, yielding the phenomenon of concept drift \cite{Gama}. We do not impose any restrictions on the concept drift. It can be real drift referring to changes of class distribution or virtual drift referring to the distribution of features. We allow sudden, incremental, gradual, and recurrent changes in the distribution of instances creating a data stream. Changes in the distribution can cause an imbalanced class system to appear in a changing configuration. 

\subsection{Incremental learning for SCM classifier}\label{sec:DSClassification:incrLearningSCM}

We assumed that the base classifier $\psi(\mathcal{L})$ wrapped by the SCM classifier is unable to learn incrementally. Consequently, an initial training set has to be used to build the classifier. This initial data set is called an initial chunk $\mathcal{B}$. The desired size of the initial bath is denoted by $|\mathcal{B}_{\mathrm{des}}|$. The initial data set is built by storing incoming examples from the data stream. By the time the initial batch is collected, the prediction is impossible. Until then, the prediction is made on the basis of \textit{a priori} probabilities estimated from the incomplete initial batch.

Since $\psi(\mathcal{L})$ is unable to learn incrementally, incremental learning is handled with changing the validation set. Incoming instances are added to the validation set until the ADWIN-based drift detector detects that the concept drift has occurred. The ADWIN-based drift detector analyses the outcomes of the corrected classifier for the instances stored in the validation set~\cite{Bifet2007}. When there is a significant difference between the older and the newer part of the validation set, the detector removes the older part of the validation set. The remaining part of the validation set is then used to correct the outcome of $\psi(\mathcal{L})$. The ADWIN-based drift detector also controls the size of the neighbourhood. Even if there is no concept drift, the detector may detect the deterioration of the classification quality when the neighbourhood becomes too large. 

The detailed procedure of the ensemble building is described in Algorithms~\ref{alg:AdwUpd} and~\ref{alg:AdwTrain}.

\begin{algorithm}[htb]
 \KwData{
  $\mathcal{V}$ -- validation set\;
  $x$ -- new instance to add\;
 }
 \KwResult{Updated validation set}
 \Begin{
    i= $\psi(\mathcal{L},\mathcal{V},x)$ \tcp*{Predict object class using corrected classifier}
    Check the prediction using ADWIN detector\;
    \uIf{ADWIN detector detects drift}{
     Ask the detector fot the newer part of the validation set $\mathcal{V}_{\mathrm{new}}$\;
     $\mathcal{V}\algAssign \mathcal{V}_{\mathrm{new}}$\;
    }
    $\mathcal{V}\algAssign \mathcal{V} \cup x$\;
 }
 \caption{Validation set update controlled by ADWIN detector.\label{alg:AdwUpd}}
\end{algorithm}

\begin{algorithm}[htb]
 \KwData{
  $x$ -- new instance;
 }
 \KwResult{Learned SCM wrapping-classifier}
 \Begin{
    \uIf{$|\mathcal{B}|\geq|\mathcal{B}_{\mathrm{des}}|$}{
        Train the SCM classifier using the procedure described in Algorithm~\ref{alg:SCM:Train} using $\mathcal{B}$ as a learning set\;
        $\mathcal{B}\algAssign \emptyset$\;
        $\mathcal{V}^{\prime}\algAssign \mathcal{V}$ \tcp*{Make a copy of the validation set}
         \ForEach{object  $x^{\prime} \in \mathcal{V}^{\prime}$}{%
            Update the validation set $\mathcal{V}$ using $x^{\prime}$ and the procedure described in Algorithm~\ref{alg:AdwUpd} 
         }
    }
    \uElseIf{Is SCM classifier trained}{
        Update the validation set $\mathcal{V}$ using $x$ and the procedure described in Algorithm~\ref{alg:AdwUpd} 
    }
    \Else{
     $\mathcal{B}\algAssign \mathcal{B} \cup x$\;
    }
 }
 \caption{Incremental learning procedure of the SCM wrapping-classifier.\label{alg:AdwTrain}}
\end{algorithm}


\section{Experimental Setup}\label{sec:ExpSetup}
To validate the classification quality obtained by the proposed approaches, the experimental evaluation, which setup is described below, is performed. 

The following base classifiers were employed:
\begin{itemize}
 \item $\psi_{\mathrm{HOE}}$ -- Hoeffding tree classifier~\cite{Pfahringer2007}
 \item $\psi_{\mathrm{NB}}$ -- Naive Bayes classifier with kernel density estimation~\cite{Hand2001}.
 \item $\psi_{\mathrm{KNN}}$ -- KNN classifier~\cite{Guo2003}.
 \item $\psi_{\mathrm{SGD}}$ -- SVM classifier built using stochastic gradient descent method~\cite{Sakr2017}.
\end{itemize}
The classifiers implemented in WEKA framework~\cite{Hall2009} were used. If not stated otherwise, the classifier parameters were set to their defaults. We have chosen the classifiers that offer both batch and incremental learning procedures. 

The experimental code was implemented using WEKA~\cite{Hall2009} framework. The source code of the algorithms is available online~\footnote{\url{https://github.com/ptrajdos/rrcBasedClassifiers/tree/develop}}~\footnote{\url{https://github.com/ptrajdos/StreamLearningPT/tree/develop}}.

During the experimental evaluation, the following classifiers  were compared: 
\begin{enumerate}
\item $\psi_{\mathrm{B}}$ -- The ADWIN-driven classifier created using the unmodified base classifier (The base classifier is able to update incrementally.)~\cite{Bifet2007}.
 \item $\psi_{\mathrm{nB}}$ -- The ADWIN-driven created using the unmodified base classifier with the incremental learning disabled. The base classifier is only retrained whenever ADWIN-based detector detects concept drift.
 \item $\psi_{\mathrm{S}}$ -- The ADWIN-driven approach using SCM correction scheme with online-learning. As described in Section~\ref{sec:DSClassification}.
 \item $\psi_{\mathrm{nS}}$ -- The ADWIN-driven approach created using SCM correction scheme but the online-learning is disabled. The SCM-corrected classifier is only retrained whenever ADWIN-based detector detects concept drift.
\end{enumerate}

To evaluate the proposed methods, the following classification-loss criteria are used~\cite{Sokolova2009}: Macro-averaged $\fdr$ (1- precision), $\fnr$ (1-recall), Matthews correlation  coefficient ($\mcc$). The Matthews coefficient is rescaled in such a way that 0 is perfect classification and 1 is the worst one. Quality measures from the macro-averaging group are considered because this kind of measures is more sensitive to the performance for minority classes. For many real-world classification problems, the minority class is the class that attracts the most attention~\cite{Leevy2018}. 

Following the recommendations of~\cite{demsar2006} and~\cite{garcia2008extension}, the statistical significance of the obtained results was assessed using the two-step procedure. The first step is to perform the Friedman test~\cite{demsar2006} for each quality criterion separately. Since multiple criteria were employed, the familywise errors (\FWER{}) should be controlled~\cite{holm1979}. To do so, the Holm~\cite{holm1979} procedure of controlling \FWER{} of the conducted Friedman tests was employed. When the Friedman test shows that there is a significant difference within the group of classifiers, the pairwise tests using the Wilcoxon signed-rank test~\cite{demsar2006} were employed. To control \FWER{} of the Wilcoxon-testing procedure, the Holm approach was employed~\cite{holm1979}. For all tests, the significance level was set to $\alpha=0.01$. 

The experiments were conducted using 48 synthetic datasets generated using the STREAM-LEARN library~\footnote{\url{https://github.com/w4k2/stream-learn}}. The properties of the datasets were as follows: Datasets size: 30k examples; Number of attributes: 8;Types of drift generated: incremental, sudden;Noise: 0\%, 10\%, 20\%; Imbalance ratio: 0 -- 4. 

Datasets used in this experiment are available online~\footnote{\url{https://github.com/ptrajdos/MLResults/blob/master/data/stream_data.tar.xz?raw=true}}

To examine the effectiveness of the incremental update algorithms, we applied an experimental procedure based on the methodology which is characteristic of data stream classification, namely, the test-then-update procedure~\cite{Gama:2010:KDD}.
The chunk size for evaluation purposes was set to 200.

\section{Results and Discussion}\label{sec:ResAndDisc}

To compare multiple algorithms on multiple benchmark sets, the average ranks approach is used. In this approach, the winning algorithm achieves a rank equal to '1', the second achieves a rank equal to '2', and so on. In the case of ties, the ranks of algorithms that achieve the same results are averaged. 

The numerical results are given in \tablenameM~\ref{table:streamHOE}~to~ \ref{table:streamSGD}. Each table is structured as follows. The first row contains the names of the investigated algorithms. Then, the table is divided into six sections -- one section is related to a single evaluation criterion. The first row of each section is the name of the quality criterion investigated in the section. The second row shows the p-value of the Friedman test. The third one shows the average ranks achieved by algorithms. The following rows show p-values resulting from the pairwise Wilcoxon test. The p-value equal to $.000$ informs that the p-values are lower than $10^{-3}$. P-values lower than $\alpha$ are bolded. Due to the page limit, the raw results are published online~\footnote{\url{https://github.com/ptrajdos/MLResults/blob/master/RandomizedClassifiers/Results_cldd_2021.tar.xz?raw=true}}

\begin{figure}[htb]
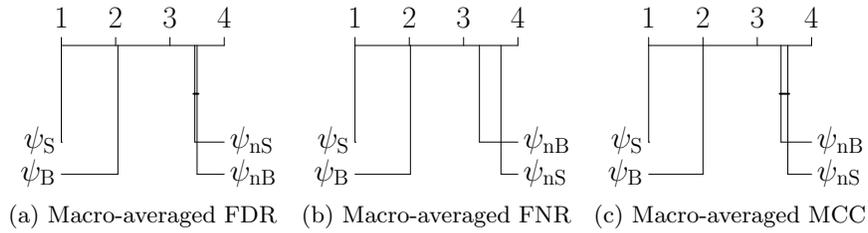

    \centering
        \subfloat[Macro-averaged $\fdr$]{\includegraphics[width=0.32\columnwidth,page=1]{\ptFiguresDirectory{sgdRank}}}
        \subfloat[Macro-averaged $\fnr$]{\includegraphics[width=0.32\columnwidth,page=2]{\ptFiguresDirectory{sgdRank}}}
        \subfloat[Macro-averaged $\mcc$]{\includegraphics[width=0.32\columnwidth,page=3]{\ptFiguresDirectory{sgdRank}}}
    \caption{Ranking plot for the stream classifiers based on $\psi_{\mathrm{SGD}}$ classifier.\label{fig:sgdRank}}
\end{figure}

To provide a visualization of the average ranks and the outcome of the statistical tests, the rank plots are used.  The rank plots are compatible with the rank plots described in~\cite{demsar2006}. That is, each classifier is placed along the line representing the values of the achieved average ranks. The classifiers between which there are no significant differences (in terms of the pairwise Wilcoxon test) are connected with a horizontal bar placed below the axis representing the average ranks. The results are visualised on figures~\ref{fig:hoeRank}~--~\ref{fig:sgdRank}. 


Let us begin with an analysis of the correction ability of the SCM approach when incremental learning is disabled. Although this kind of analysis has been already done~\cite{AMCS,CBM}, in this work it should be done again since the definition of the neighbourhood is significantly changed (see Section~\ref{sec:ClassifierCorrection:SCM}).  To assess the impact of the SCM-based correction, we compare the algorithms $\psi_{\mathrm{nB}}$ and $\psi_{\mathrm{nS}}$ for different base classifiers. For $\psi_{\mathrm{HOE}}$ and $\psi_{\mathrm{NB}}$ base classifiers the employment of SCM-based correction allows achieving significant improvement in terms of all quality criteria (see Figures~\ref{fig:hoeRank} and~\ref{fig:nbRank}). For the remaining base classifiers, on the other hand, there are no significant differences between $\psi_{\mathrm{nB}}$ and $\psi_{\mathrm{nS}}$. These results confirm observations previously made in~\cite{AMCS,CBM}. That is, the correction ability of the SCM approach is more noticeable for classifiers that are considered to be weaker ones. The previously observed correction ability holds although the extensive grid-search technique is not applied.

In this paper, the SCM-based approach is proposed to be used as a wrapping-classifier that handles the incremental learning for base classifiers that are unable to be updated incrementally. Consequently, now we are going to analyse the SCM approach in that scenario. The results show that $\psi_{\mathrm{S}}$ significantly outperforms $\psi_{\mathrm{nB}}$ for all base classifiers and quality criteria. It means that it works great as the incremental-learning-handling wrapping-classifier. What is more, it outperforms $\psi_{\mathrm{nS}}$ also for all base classifiers and criteria. It clearly shows that the source of the achieved improvement does not lie in the batch-learning-improvement-ability but the ability to handle incremental learning is also present. Moreover, it handles incremental learning more effective than the base classifiers designed to do so. This observation is confirmed by the fact that $\psi_{\mathrm{S}}$ also outperforms $\psi_{\mathrm{B}}$ for all base classifiers and quality criteria.


{
\setlength\tabcolsep{2.0pt}%

\begin{table}[htb]
\centering\scriptsize
\caption{Statistical evaluation for the stream classifiers based on $\psi_{\mathrm{HOE}}$ classifier.\label{table:streamHOE}}
\begin{tabular}{c|cccc|cccc|cccc}
  & $\psi_{\mathrm{B}}$ & $\psi_{\mathrm{nB}}$ & $\psi_{\mathrm{S}}$ & $\psi_{\mathrm{nS}}$ & $\psi_{\mathrm{B}}$ & $\psi_{\mathrm{nB}}$ & $\psi_{\mathrm{S}}$ & $\psi_{\mathrm{nS}}$ & $\psi_{\mathrm{B}}$ & $\psi_{\mathrm{nB}}$ & $\psi_{\mathrm{S}}$ & $\psi_{\mathrm{nS}}$ \\ 
  \midrule
Crit. Name&\multicolumn{4}{c|}{MaFDR}&\multicolumn{4}{c|}{MaFNR}&\multicolumn{4}{c}{MaMCC}\\
Friedman p-value&\multicolumn{4}{c|}{\textbf{1.213e-28}}&\multicolumn{4}{c|}{\textbf{5.963e-28}}&\multicolumn{4}{c}{\textbf{5.963e-28}}\\
Average Rank& 2.000 & 3.812 & 1.00 & 3.188 & 2.000 & 3.583 & 1.00 & 3.417 & 2.000 & 3.667 & 1.00 & 3.333 \\ 
   \midrule
  $\psi_{\mathrm{B}}$ &  & \textbf{.000} & \textbf{.000} & \textbf{.000} &  & \textbf{.000} & \textbf{.000} & \textbf{.000} &  & \textbf{.000} & \textbf{.000} & \textbf{.000} \\ 
  $\psi_{\mathrm{nB}}$ &  &  & \textbf{.000} & \textbf{.000} &  &  & \textbf{.000} & .111 &  &  & \textbf{.000} & \textbf{.002} \\ 
  $\psi_{\mathrm{S}}$ &  &  &  & \textbf{.000} &  &  &  & \textbf{.000} &  &  &  & \textbf{.000} \\ 
  \end{tabular}
\end{table}
}

{
\setlength\tabcolsep{2.0pt}%
\def\arraystretch{0.9}%

\begin{table*}[htb]
\centering\scriptsize
\caption{Statistical evaluation for the stream classifiers based on $\psi_{\mathrm{NB}}$ classifier.\label{table:streamNB}}
\begin{tabular}{c|cccc|cccc|cccc}
  & $\psi_{\mathrm{B}}$ & $\psi_{\mathrm{nB}}$ & $\psi_{\mathrm{S}}$ & $\psi_{\mathrm{nS}}$ & $\psi_{\mathrm{B}}$ & $\psi_{\mathrm{nB}}$ & $\psi_{\mathrm{S}}$ & $\psi_{\mathrm{nS}}$ & $\psi_{\mathrm{B}}$ & $\psi_{\mathrm{nB}}$ & $\psi_{\mathrm{S}}$ & $\psi_{\mathrm{nS}}$ \\ 
  \midrule
Crit. Name&\multicolumn{4}{c|}{MaFDR}&\multicolumn{4}{c|}{MaFNR}&\multicolumn{4}{c}{MaMCC}\\
Friedman p-value&\multicolumn{4}{c|}{\textbf{3.329e-28}}&\multicolumn{4}{c|}{\textbf{3.329e-28}}&\multicolumn{4}{c}{\textbf{1.739e-28}}\\
Average Rank& 2.021 & 3.771 & 1.00 & 3.208 & 2.000 & 3.708 & 1.00 & 3.292 & 2.000 & 3.792 & 1.00 & 3.208 \\ 
   \midrule
  $\psi_{\mathrm{B}}$ &  & \textbf{.000} & \textbf{.000} & \textbf{.000} &  & \textbf{.000} & \textbf{.000} & \textbf{.000} &  & \textbf{.000} & \textbf{.000} & \textbf{.000} \\ 
  $\psi_{\mathrm{nB}}$ &  &  & \textbf{.000} & \textbf{.000} &  &  & \textbf{.000} & \textbf{.001} &  &  & \textbf{.000} & \textbf{.000} \\ 
  $\psi_{\mathrm{S}}$ &  &  &  & \textbf{.000} &  &  &  & \textbf{.000} &  &  &  & \textbf{.000} \\ 
  \end{tabular}
\end{table*}
}

{
\setlength\tabcolsep{2.0pt}%
\def\arraystretch{0.9}%

\begin{table*}[htb]
\centering\scriptsize
\caption{Statistical evaluation for the stream classifiers based on $\psi_{\mathrm{KNN}}$ classifier.\label{table:streamKNN}}
\begin{tabular}{c|cccc|cccc|cccc}
  & $\psi_{\mathrm{B}}$ & $\psi_{\mathrm{nB}}$ & $\psi_{\mathrm{S}}$ & $\psi_{\mathrm{nS}}$ & $\psi_{\mathrm{B}}$ & $\psi_{\mathrm{nB}}$ & $\psi_{\mathrm{S}}$ & $\psi_{\mathrm{nS}}$ & $\psi_{\mathrm{B}}$ & $\psi_{\mathrm{nB}}$ & $\psi_{\mathrm{S}}$ & $\psi_{\mathrm{nS}}$ \\ 
  \midrule
Crit. Name&\multicolumn{4}{c|}{MaFDR}&\multicolumn{4}{c|}{MaFNR}&\multicolumn{4}{c}{MaMCC}\\
Friedman p-value&\multicolumn{4}{c|}{\textbf{1.883e-27}}&\multicolumn{4}{c|}{\textbf{1.883e-27}}&\multicolumn{4}{c}{\textbf{1.883e-27}}\\
Average Rank& 2.000 & 3.521 & 1.00 & 3.479 & 2.000 & 3.542 & 1.00 & 3.458 & 2.000 & 3.500 & 1.00 & 3.500 \\ 
   \midrule
$\psi_{\mathrm{B}}$ &  & \textbf{.000} & \textbf{.000} & \textbf{.000} &  & \textbf{.000} & \textbf{.000} & \textbf{.000} &  & \textbf{.000} & \textbf{.000} & \textbf{.000} \\ 
  $\psi_{\mathrm{nB}}$ &  &  & \textbf{.000} & .955 &  &  & \textbf{.000} & .545 &  &  & \textbf{.000} & .757 \\ 
  $\psi_{\mathrm{S}}$ &  &  &  & \textbf{.000} &  &  &  & \textbf{.000} &  &  &  & \textbf{.000} \\ 
  \end{tabular}
\end{table*}
}

{
\setlength\tabcolsep{2.0pt}%
\def\arraystretch{0.9}%

\begin{table*}[htb]
\centering\scriptsize
\caption{Statistical evaluation for the stream classifiers based on $\psi_{\mathrm{SGD}}$ classifier.\label{table:streamSGD}}
\begin{tabular}{c|cccc|cccc|cccc}
  & $\psi_{\mathrm{B}}$ & $\psi_{\mathrm{nB}}$ & $\psi_{\mathrm{S}}$ & $\psi_{\mathrm{nS}}$ & $\psi_{\mathrm{B}}$ & $\psi_{\mathrm{nB}}$ & $\psi_{\mathrm{S}}$ & $\psi_{\mathrm{nS}}$ & $\psi_{\mathrm{B}}$ & $\psi_{\mathrm{nB}}$ & $\psi_{\mathrm{S}}$ & $\psi_{\mathrm{nS}}$ \\ 
  \midrule
Crit. Name&\multicolumn{4}{c|}{MaFDR}&\multicolumn{4}{c|}{MaFNR}&\multicolumn{4}{c}{MaMCC}\\
Friedman p-value&\multicolumn{4}{c|}{\textbf{3.745e-27}}&\multicolumn{4}{c|}{\textbf{1.563e-27}}&\multicolumn{4}{c}{\textbf{1.563e-27}}\\
Average Rank& 2.042 & 3.500 & 1.00 & 3.458 & 2.021 & 3.292 & 1.00 & 3.688 & 2.000 & 3.438 & 1.00 & 3.562 \\ 
   \midrule
$\psi_{\mathrm{B}}$ &  & \textbf{.000} & \textbf{.000} & \textbf{.000} &  & \textbf{.000} & \textbf{.000} & \textbf{.000} &  & \textbf{.000} & \textbf{.000} & \textbf{.000} \\ 
  $\psi_{\mathrm{nB}}$ &  &  & \textbf{.000} & .947 &  &  & \textbf{.000} & \textbf{.005} &  &  & \textbf{.000} & .088 \\ 
  $\psi_{\mathrm{S}}$ &  &  &  & \textbf{.000} &  &  &  & \textbf{.000} &  &  &  & \textbf{.000} \\ 
  \end{tabular}
\end{table*}
}


\begin{figure}[htb]
    \centering
        \subfloat[Macro-averaged $\fdr$]{\includegraphics[width=0.32\columnwidth,page=1]{\ptFiguresDirectory{hoeRank}}}
        \subfloat[Macro-averaged $\fnr$]{\includegraphics[width=0.32\columnwidth,page=2]{\ptFiguresDirectory{hoeRank}}}
        \subfloat[Macro-averaged $\mcc$]{\includegraphics[width=0.32\columnwidth,page=3]{\ptFiguresDirectory{hoeRank}}}
    \caption{Ranking plot for the stream classifiers based on $\psi_{\mathrm{HOE}}$ classifier.\label{fig:hoeRank}}
\end{figure}

\begin{figure}[htb]
    \centering
        \subfloat[Macro-averaged $\fdr$]{\includegraphics[width=0.32\columnwidth,page=1]{\ptFiguresDirectory{nbRank}}}
        \subfloat[Macro-averaged $\fnr$]{\includegraphics[width=0.32\columnwidth,page=2]{\ptFiguresDirectory{nbRank}}}
        \subfloat[Macro-averaged $\mcc$]{\includegraphics[width=0.32\columnwidth,page=3]{\ptFiguresDirectory{nbRank}}}
    \caption{Ranking plot for the stream classifiers based on $\psi_{\mathrm{NB}}$ classifier.\label{fig:nbRank}}
\end{figure}

\begin{figure}[htb]
    \centering
        \subfloat[Macro-averaged $\fdr$]{\includegraphics[width=0.32\columnwidth,page=1]{\ptFiguresDirectory{knnRank}}}
        \subfloat[Macro-averaged $\fnr$]{\includegraphics[width=0.32\columnwidth,page=2]{\ptFiguresDirectory{knnRank}}}
        \subfloat[Macro-averaged $\mcc$]{\includegraphics[width=0.32\columnwidth,page=3]{\ptFiguresDirectory{knnRank}}}
    \caption{Ranking plot for the stream classifiers based on $\psi_{\mathrm{KNN}}$ classifier.\label{fig:knnRank}}
\end{figure}

\section{Conclusions}\label{sec:Conclusions}

In this paper, we propose a modified SCM classifier to be used as a wrapping-classifier that allows incremental learning of classifiers that are not designed to be incrementally updated. We applied two modifications of the SCM wrapping-classifier originally described in~\cite{AMCS,CBM}.  The first one is a modified neighbourhood definition. The newly proposed neighbourhood does not need an excessive grid-search procedure to be performed to find the best set of parameters. Due to the modified neighbourhood definition, the computational cost of performing the SCM-based correction is significantly smaller. The second modification is to incorporate ADWIN-based approach to create and manage the validation set used by SCM-based algorithm. This modification not only allows the proposed method to effectively deal with the concept drift but also it can shrink the neighbourhood when it becomes too wide. 

The experimental results show that the proposed approach outperforms the reference methods for all investigated base classifiers in terms of all considered quality criteria. 

The results obtained in this study are very promising. Consequently, we are going to continue our research related to the employment of randomised classifiers in the task of stream learning. Our next step will probably be a proposition of a stream learning ensemble that used the SCM-correction method proposed in this paper.

\subsubsection*{Acknowledgments.} This work was supported by the statutory funds of the Department of Systems and Computer Networks, Wroclaw University of Science and Technology.
\FloatBarrier

\bibliography{bibliography}

\end{document}